\documentclass[journal]{IEEEtran}

\usepackage[hidelinks]{hyperref}

\usepackage{amsmath,amssymb,amsfonts}
\usepackage{graphicx}
\usepackage{dblfloatfix}
\usepackage{booktabs}
\usepackage{siunitx}
\usepackage{xcolor,colortbl}
\usepackage{bm,bbm}
\usepackage{orcidlink}
\usepackage{url}
\usepackage{nicematrix}
\usepackage{multirow}
\usepackage{enumitem}
\usepackage{amssymb}

\usepackage{cite}

\definecolor{best}{RGB}{255,180,180}    
\definecolor{second}{RGB}{255,215,160}  
\definecolor{third}{RGB}{255,245,170}   

\def\highres{0} 

\begin{document}

\title{WaveDINO: Learning-Based Atmospheric Correction of Unwrapped InSAR Interferograms Validated by GNSS: Results at Laguna del Maule and Campi Flegrei Volcanoes}

\author{
Robert~Popescu\orcidlink{0009-0007-8124-737X},
Juliet~Biggs\orcidlink{0000-0002-4855-039X},
Tianyuan~Zhu\orcidlink{0009-0007-9926-5648},
and Nantheera~Anantrasirichai\orcidlink{0000-0002-2122-5781}
\thanks{This work is supported by the European Research Council (ERC) under the European Union’s Horizon 2020 research and innovation programme (MAST; Grant No.\ 101003173) and the NERC Centre for the Observation and Modelling of Earthquakes, Volcanoes, and Tectonics (COMET, \url{http://comet.nerc.ac.uk}), a partnership between UK Universities and the British Geological Survey.}%
\thanks{R. Popescu, J. Biggs, T. Zhu and N. Anantrasirichai are with the University of Bristol, Bristol, U.K.
(e-mail: zm18440@bristol.ac.uk; Juliet.Biggs@bristol.ac.uk; tianyuan.zhu@bristol.ac.uk; N.Anantrasirichai@bristol.ac.uk).}%
}


\maketitle

\begin{abstract}
Interferometric Synthetic Aperture Radar (InSAR) enables effective monitoring of volcanic deformation; however, the observed signals are often corrupted by atmospheric phase delays, seasonal surface changes, and decorrelation effects. Existing atmospheric correction methods, such as numerical weather model–based methods, can reduce these effects but do not consistently remove atmospheric artefacts and may introduce residual biases. To address these limitations, we propose a novel learning-based method for denoising unwrapped InSAR interferograms, using a hybrid training strategy that combines physically motivated synthetic deformation with real atmospheric noise. Specifically, we introduce \emph{WaveDINO}, a wavelet-based multi-scale denoising framework conditioned on frozen DINOv3 foundation-model features and terrain information.  Training uses synthetic magma-source deformation superimposed on short-term interferograms to expose the network to realistic atmospheric statistics while retaining known ground truth. Performance is evaluated on both controlled synthetic data and long-term real interferograms from Laguna del Maule (Chile) and Campi Flegrei (Italy), with independent GNSS measurements used for validation. WaveDINO consistently outperforms competing models, improving agreement with GNSS measurements, and reducing mean GNSS misfit by approximately 3\% and 19\% at two sites, respectively, while surpassing weather-model-based corrections.

\end{abstract}


\section{Introduction}

More than 8\% of the world’s population lives within 100~km of a volcano with at least one significant eruption \cite{Freire2019}, underscoring the need for robust volcano monitoring and eruption forecasting \cite{Poland2020, sparks2012monitoring}. Spaceborne Interferometric Synthetic Aperture Radar (InSAR) enables wide-area, repeat observations of ground displacement by exploiting phase differences between repeated radar acquisitions to form interferograms. These measurements are critical for tracking volcanic deformation—changes in surface shape or elevation driven by subsurface magma movement or other geophysical processes—which often precede or accompany unrest and eruptions \cite{Biggs2014, Pritchard2022, BIGGS2026115377}.

However, interferograms encode not only deformation but also atmospheric path delays (e.g., stratified and turbulent tropospheric components), which can mimic, mask, or distort geophysical signals and degrade both automatic detection and geophysical inversions. Numerical weather model–based corrections, such as GACOS \cite{gacos1, gacos2, gacos3}, can reduce atmospheric artefacts, yet they rely on dense spatiotemporal sampling and the fidelity of model fields; in practice, residuals can remain and, in some cases, spurious artefacts may be introduced \cite{albino:automated:2020, kirsten2020}. GNSS-based corrections offer an independent path-delay constraint but are limited by the sparse, uneven distribution of ground stations around many volcanoes \cite{Albino2025GNSS}. Consequently, there is a pressing need for methods that selectively suppress atmospheric noise while preserving true deformation.

Recent learning-based image restoration has achieved strong results across denoising and related low-level vision tasks, including U-Net–style encoder–decoders, residual networks, attention mechanisms, and, more recently, transformer-based models \cite{Zhang2018, dudhane2022burst, Zamir2020MIRNet, Zamir2021MPRNet, zhang2018residual, zhang2020rdnir, Wang_2022_CVPR, Yue_2020, Zhang2021, Cho2021, Abdullah2020, Jiwon2016, Cavigelli2017, Liu2019, Gu2019, Niu2020, chen2020pre, cao2021vsrt, yang2020learning}. Yet most denoisers are trained on natural-image datasets and assume pixel-wise independent and identically distributed random noise (e.g., Gaussian/Poisson), whereas InSAR atmospheric delay is spatially coherent, nonstationary, and mesoscale, often exhibiting smooth, low-wavenumber patterns that overlap spectrally with long-wavelength deformation. As a result, off-the-shelf denoisers tend to oversmooth or remove true deformation, induce ringing near phase discontinuities, and fail under decorrelation, layover, and coherence loss typical of InSAR images. While several studies have retrained CNN-based models on synthetic atmospheric realisations for InSAR denoising \cite{Vijay2023, BRENCHER2025, zhang2023}, open questions persist regarding generalisation to real interferograms (sensor-, site-, and orbit-specific statistics), robustness to phase unwrapping errors (e.g., branch cuts, residue clustering) and topographic residuals (DEM errors), and validation against independent ground truth. Traditional corrections thus remain widely used \cite{Jolivet2011, gacos1, gacos2, gacos3, Zebker2021}, despite their  limitations.

To address this challenge, we propose a denoising framework guided by two complementary principles: (i) explicitly modelling the multi-scale spatial structure of atmospheric phase screens, and (ii) injecting global contextual information learned from large-scale self-supervised pretraining to improve robustness under real interferometric artefacts. Our backbone builds on a wavelet-based encoder--decoder design \cite{zou2024wavemamba}, where the input is decomposed into multi-resolution subbands so that broad, spatially coherent atmospheric components can be modelled primarily in the low-frequency pathway while fine deformation-related structure is preserved through high-frequency skip connections.

In addition, we condition the denoiser on dense features extracted from a frozen DINOv3 foundation backbone (ConvNeXt-Base; \texttt{dinov3-convnext-base}) via cross-attention at multiple scales. DINOv3 provides high-quality dense representations that transfer well across tasks and data distributions without fine-tuning, making it a strong source of global context that complements the locality of convolutional processing \cite{Simeoni2025dinov3}. In our setting, this global context helps disambiguate large-scale atmospheric patterns from true deformation and improves stability under decorrelation and spatially nonstationary noise.

Our design targets three key gaps in current InSAR denoising practice: (1) supervised learning is limited by the lack of ground-truth deformation for real interferograms; (2) dominant nuisance terms are spatially correlated and nonstationary, violating the independent and identically distributed (i.i.d.) noise assumptions of many off-the-shelf denoisers; and (3) models trained on simplified synthetic noise often generalize poorly to real interferograms and are rarely validated against independent geodetic observations. We address these gaps by training with physically motivated synthetic deformation superimposed on real interferograms to expose the network to realistic atmospheric statistics, and by combining wavelet multi-scale processing with DINOv3 conditioning to better suppress atmospheric artefacts while preserving deformation structure, validated against GNSS at two contrasting volcanic systems.

We refer to our method as \textit{WaveDINO}, a wavelet multi-scale denoiser conditioned on pretrained DINOv3 ConvNeXt-Base features with DEM conditioning via Feature-wise Linear Modulation (FiLM) \cite{perez2018film} for InSAR atmospheric phase correction.

Our main contributions are:
\begin{enumerate}[leftmargin=0pt, labelindent=0pt, itemindent=22pt, listparindent=0pt]
    \item \textbf{WaveDINO} is a wavelet-based multi-scale denoising framework for unwrapped InSAR, conditioned on DINOv3 features and terrain information. It explicitly targets spatially coherent, nonstationary atmospheric artefacts while preserving deformation-related gradients and localised signals.
    
    \item \textbf{DINOv3 Cross-Attention (DCA)} integrates DINOv3 features with model representations through cross-attention, enabling effective fusion of global semantic context with low-frequency structure.
    
    \item \textbf{Terrain Modulation Block (TMB)} provides terrain-aware conditioning by aligning DEM-derived features with the network representation, improving robustness to terrain-dependent atmospheric variability.
    
    \item \textbf{Phase Ramp Head (PRH)} models large-scale atmospheric artefacts as a global planar trend, enabling efficient removal of low-order biases and allowing the backbone to focus on residual structure.
    
    \item \textbf{Hybrid training strategy} constructs training samples by superimposing physically motivated synthetic deformation onto short-duration real interferograms, enabling supervised learning without ground truth while exposing the model to realistic atmospheric statistics and improving robustness over purely synthetic training.

\end{enumerate}

Overall, our results indicate that learning-based denoising trained on hybrid (synthetic+real) data can effectively attenuate atmospheric artefacts while preserving deformation signatures, providing a complementary pathway to traditional atmospheric corrections and supporting more reliable volcano monitoring from InSAR.

\section{Datasets and Study Areas}

The InSAR data used in this paper was acquired by the Sentinel-1 radar mission operated by the European Space Agency (ESA). The data was processed using LiCSAR \cite{lazecky2020}, an automated InSAR processing system developed by the Centre for Observation and Modelling of Earthquakes, Volcanoes, and Tectonics (COMET). We use reference areas for the interferograms obtained from the time series generated automatically by LICSBAS \cite{Licsbas} for the COMET Volcano Portal (\url{https://comet.nerc.ac.uk/comet-volcano-portal/}). 

For our case studies, we selected two caldera volcanoes with ongoing deformation but contrasting atmospheric and environmental conditions. Laguna del Maule (Chile) is located in a high-altitude Andean setting with strong topographic relief and highly variable meteorological conditions \cite{Feigl2013}, while Campi Flegrei (Italy) is a densely monitored caldera situated in a low-relief, coastal and urbanised environment \cite{Bevilacqua2024}. 

\begin{figure}
    \centering
    \if\highres1            
        \includegraphics[width=1\linewidth]{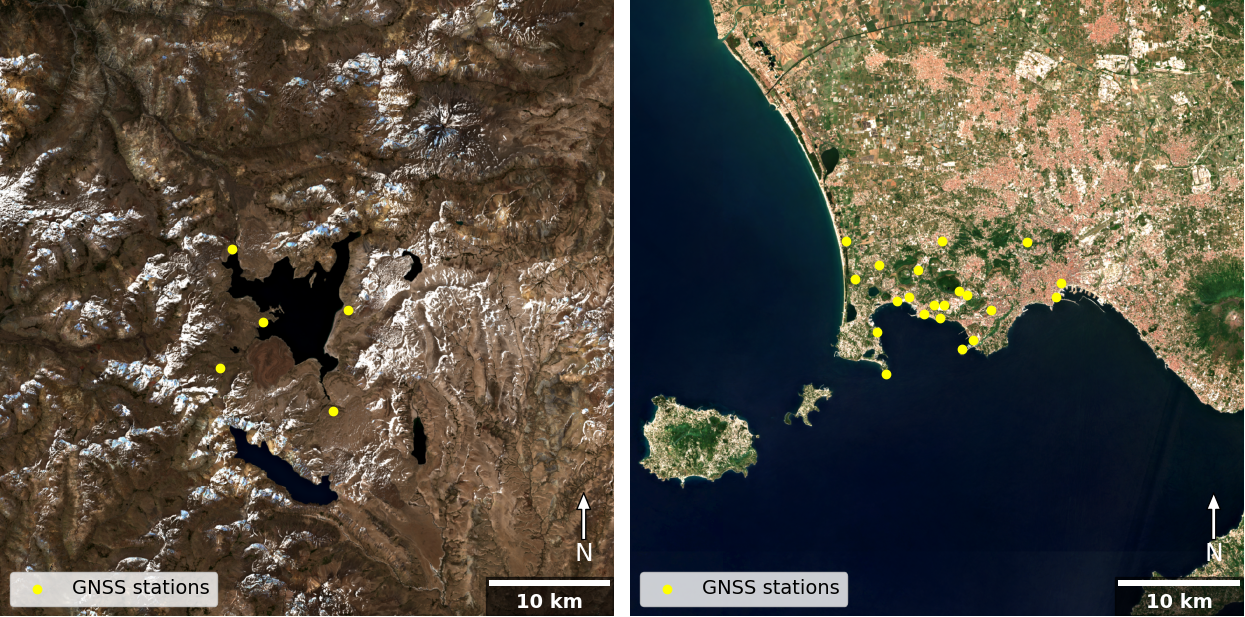}
    \else
        \includegraphics[width=1\linewidth]{images/satellite_diagram.png}
    \fi
    
    \caption{Optical satellite imagery illustrating the contrasting environmental and geological settings of the two study areas. Left: Laguna del Maule, Chile. Right: Campi Flegrei, Italy. 
    Satellite imagery from Copernicus Sentinel-2 (ESA, 2025–2026), processed using Google Earth Engine.}
    \label{fig:satellite}
\end{figure}

Figure~\ref{fig:satellite} illustrates the markedly different environmental settings of the two study areas. Laguna del Maule is characterised by rugged topography and seasonal snow cover, conditions that promote strong stratified tropospheric delays and coherence loss. In contrast, Campi Flegrei is influenced by coastal and urban boundary-layer processes, resulting in spatially heterogeneous atmospheric artefacts. These contrasting environments provide a stringent test for atmospheric correction approaches and enable evaluation across distinct atmospheric regimes.

\subsection{Sentinel-1 Acquisition and Processing}
\label{sec:s1_processing}

We use Sentinel-1 IW mode data processed automatically by the LiCSAR pipeline \cite{lazecky2020} and time-series referencing via LICSBAS \cite{Licsbas}. No additional pre-processing beyond LiCSAR and LICSBAS defaults was performed. Two study areas are investigated in this study: (i) Laguna del Maule, which demonstrates large deformation signals; and (ii) Campi Flegrei, which represents the smaller-signal case. Specifically:

\begin{itemize}[leftmargin=*]
   \item Laguna del Maule (LdM, $70.492^\circ W$, $36.058^\circ S$): tracks \textbf{18A} and \textbf{83D}, 622 epochs from 06-10-2014 to 26-08-2024.
   \item Campi Flegrei (CF, $14.139^\circ E$, $40.827^\circ N$): tracks \textbf{22D}, \textbf{44A}, and \textbf{124D}, 1008 epochs from 07-10-2014 to 22-09-2025.
 \end{itemize}

LiCSAR handles precise orbits, co-registration, TOPS (Terrain Observation with Progressive Scans) burst stitching, interferogram formation, multilooking, and geocoding, while LICSBAS provides the time-series reference point used for offset removal in Section~\ref{sec:preproc}.\footnote{We follow the LiCSAR/LICSBAS defaults; detailed parameterisation (e.g., multilook factors, DEM source, and unwrapping options) can be retrieved from the LiCSAR metadata supplied with each frame.} We analyse a clipped region of interest of 50$\times$50~km centred on each volcano, with an image size of 500$\times$500 pixels at 50~m pixel spacing. We apply weather-model-based atmospheric corrections using GACOS \cite{gacos1} and test our models using both corrected and uncorrected interferograms. 

\subsubsection{Laguna del Maule}

Laguna del Maule is a silicic volcanic system located in the southern Andes along the Chile–Argentina border and is one of the most rapidly and consistently deforming volcanoes in the world \cite{Feigl2013}. The system comprises a caldera complex surrounded by numerous rhyolitic domes and lava flows. Satellite geodetic observations over the past two decades have revealed persistent uplift at rates exceeding several centimetres per year, suggesting significant magma accumulation at shallow depths beneath the caldera \cite{Feigl2013,LeMevel2015}. Despite the absence of recent eruptive activity, the magnitude and duration of deformation indicate that Laguna del Maule is an active magmatic system undergoing notable subsurface changes \cite{Feigl2013}.

\subsubsection{Campi Flegrei}

Campi Flegrei is a large caldera system located west of Naples, Italy, and is one of the most hazardous and densely populated volcanic regions globally. The caldera has experienced multiple eruptive phases over the last 40,000 years, including the Campanian Ignimbrite and Neapolitan Yellow Tuff eruptions \cite{Bevilacqua2024}. Today, Campi Flegrei is characterised by persistent hydrothermal and magmatic unrest expressed through seismic activity, ground deformation, and gas emissions, collectively known as bradyseism. Uplift episodes recorded since the mid-20th century have been interpreted as evidence of magma or gas pressurisation at shallow depths, prompting ongoing monitoring efforts to assess eruption potential and improve hazard forecasting \cite{Bevilacqua2024}. 

\subsection{Synthetic data}
\label{sec:synthetic_data}

\begin{figure}
    \centering
    \if\highres1            
        \includegraphics[width=1\linewidth]{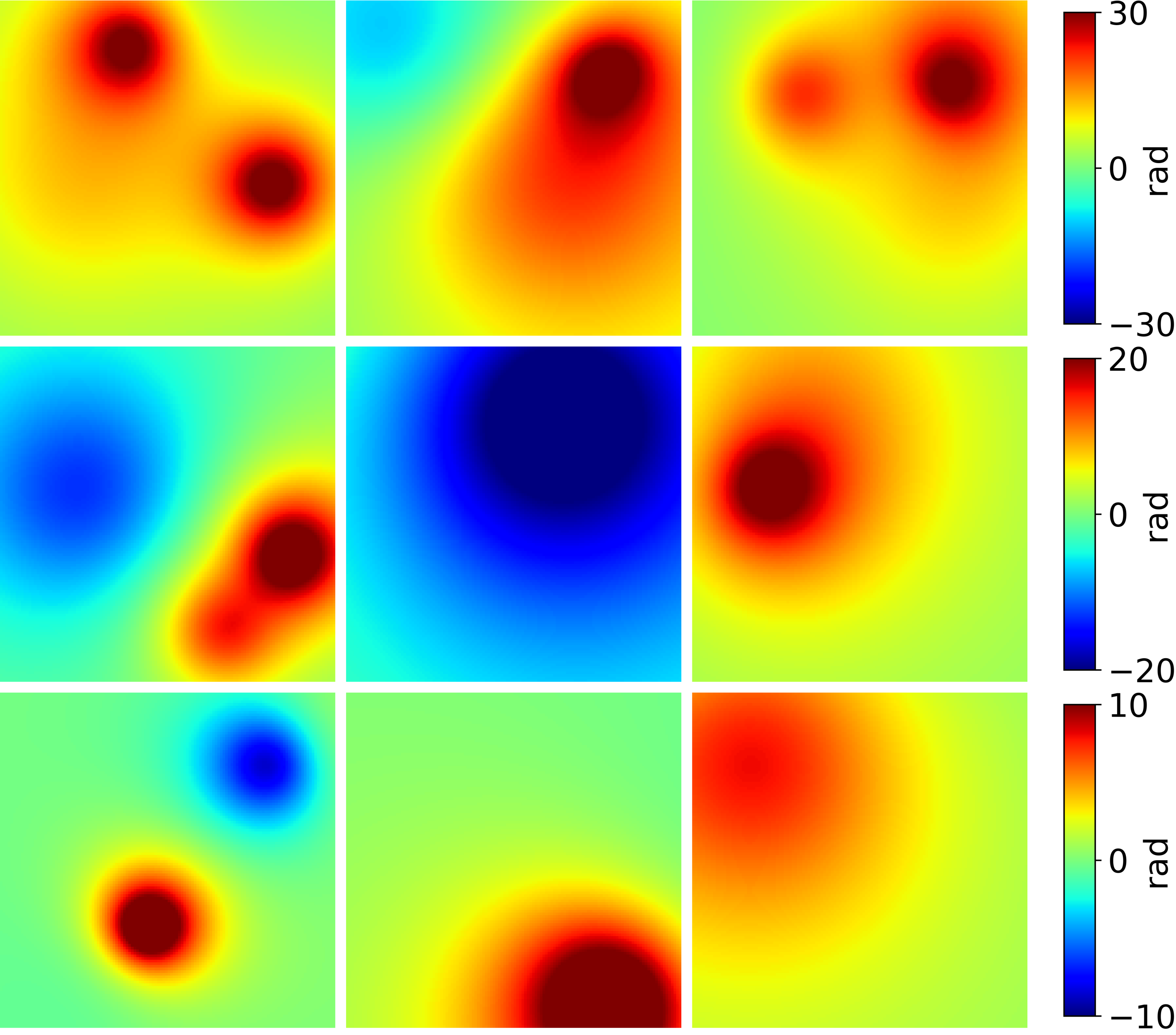}
    \else
        \includegraphics[width=1\linewidth]{images/deform_examples.png}
    \fi
    
    \caption{Examples of synthetic deformation used during training. Each panel shows the superposition of 1 to 4 randomly shifted magma-source deformations.
    Rows differ only in deformation magnitude (top: $\pm 30$ rad, middle: $\pm 20$ rad, bottom: $\pm 10$ rad), as indicated by the colour bars; columns show different random realisations.
    }
    \label{deform_examples}
\end{figure}

\begin{figure*}
    \centering
    \if\highres1            
        \includegraphics[width=1\linewidth]{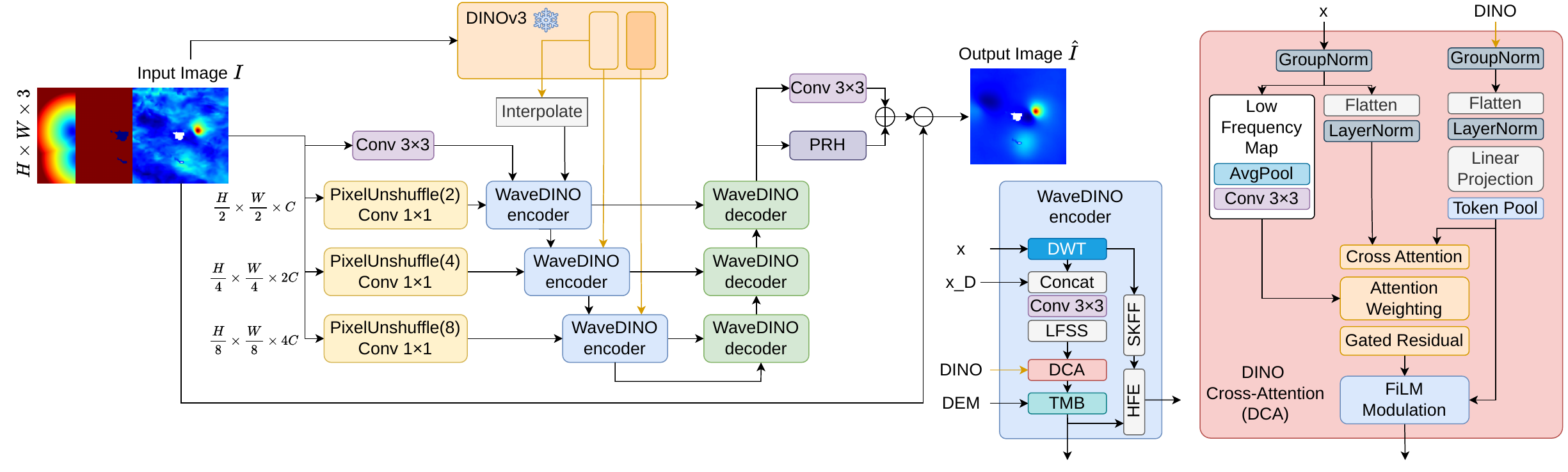}
    \else
        \includegraphics[width=1\linewidth, trim=0 2mm 0 0mm, clip]{images/diagram_1.pdf}
    \fi
    
    \caption{Diagram of the proposed WaveDINO framework for atmospheric phase estimation and removal in unwrapped InSAR. A wavelet-based encoder–decoder models multi-scale structure, enhanced by cross-attention with DINOv3 features and DEM-based FiLM conditioning. LFSS: Low-Frequency State Space; SKFF: Selective Kernel Feature Fusion; HFE: High-Frequency Enhancement.}
    \label{fig:diagram_big}
\end{figure*}

Because ground-truth deformation fields are unavailable for real interferograms, supervised learning on purely observed data is not feasible. We therefore simulate deformation patterns using pressurising magma sources and use these as training targets. To increase diversity and task complexity, each sample combines 1–4 sources with random translations to avoid overlap (see \autoref{deform_examples}). This multi-source setup exposes the network to a broader range of deformation morphologies and encourages more accurate reconstruction beyond simple single-lobe patterns.

We generated synthetic deformation $\phi_{\text{syn}}$ using a point source model \cite{Mogi1958}. The simulation parameters were sampled as follows: 36 incidence angles ($^\circ$) from 0--90, 28 satellite heading angles ($^\circ$) from 4--350, 19 log$_{10}$ volume-change parameters $\log_{10}\Delta V$ from 5--7.9 (i.e., $\Delta V$=$10^{\log_{10}\Delta V}$ m$^3$), and 18 source depths from 1.5--8 km. To maintain physically plausible scaling, depth was coupled to volume such that smaller volumes were paired with shallower sources and larger volumes with deeper sources. The incidence and heading angles were sampled over a wide range to span diverse look geometries.

To incorporate realistic atmospheric artefacts during training while preserving a known target, we form the network input by simple addition of synthetic deformation and short-duration interferograms: $x = \phi_{\text{syn}} + \phi_{\text{atm}}$, where $\phi_{\text{atm}}$ represents real atmospheric and other nuisance phase contributions. We use short-duration interferograms ($\leq$18 days) because deformation evolves slowly relative to these short baselines \cite{Feigl2013, Bevilacqua2024}, so such interferograms contain negligible deformation and can be treated as predominantly atmospheric and decorrelation noise. Long-term interferograms ($>$18 days) are dominated by volcano deformation \cite{LeMével2021, Polcari2022}, which is located roughly in the middle of the image, meaning that we can discard the middle and include the edges in the dataset.


\subsection{Global Navigation Satellite System (GNSS) measurements}

Ground-based GNSS instruments can track three-dimensional movements of the ground surface, typically providing millimetric precision at daily resolution. In this work, we use them to evaluate the geophysical consistency of the denoised results. We focus on long temporal baselines ($>$18 days), as deformation accumulates over time and becomes more distinguishable relative to tropospheric variability, whereas short-term interferograms are typically dominated by atmospheric effects, giving a low signal-to-noise ratio \cite{Licsbas}. For each station, we project the three-dimensional measurements into the satellite line-of-sight for each track. GNSS time series can be noisy, particularly in the vertical component; therefore, we apply a 1-D interpolation using \texttt{make\_splrep} from \texttt{scipy}. This creates a smoothing B-spline function with bounded error, minimising derivative jumps in the time series.

For Laguna del Maule, we use 5 GNSS stations (COLO, LDMP, MAU2, NIEB, and PUEL) from the OVDAS–SERNAGEOMIN deformation monitoring network, using daily time series spanning 2014-01-01 to 2023-12-30.
For Campi Flegrei, ground deformation is monitored by the Instituto Nazionale di Geofisica e Vulcanologia–Osservatorio Vesuviano (INGV-OV) continuous GNSS network. In this study, we use weekly time series from 21 GNSS stations spanning January 2000 to October 2023 \cite{DeMartino2021, DeMartinoDataset2023}; two stations (CMIS and NAMM) are excluded from the GNSS–InSAR comparison because the surrounding InSAR pixels are frequently decorrelated/masked (low coherence), preventing reliable sampling.


\section{Methodology}

\subsection{Network Architecture}

A diagram of the proposed framework is shown in Figure~\ref{fig:diagram_big}.
We introduce an InSAR-oriented adaptation of WaveMamba~\cite{zou2024wavemamba}, exploiting its wavelet-based multi-scale representation and convolutional feature extraction to mitigate atmospheric phase artefacts. The architecture is further refined to better accommodate spatially correlated signals by incorporating complementary priors from a foundation model alongside terrain-related information.

To improve feature expressiveness, we progressively increase the number of feature channels with depth. The encoder channels follow the progression:
$32 \rightarrow 64 \rightarrow 128$,
while the decoder mirrors this structure in reverse:
$128 \rightarrow 64 \rightarrow 32$.

This multi-scale design allows deeper layers to operate at lower spatial resolution but higher channel dimensionality, enabling the network to model large-scale spatial correlations that frequently characterise atmospheric artefacts. Shallower layers retain higher spatial resolution to preserve deformation-related spatial detail in the unwrapped phase field, such as sharp gradients, localised anomalies, and fine structural boundaries. The input is decomposed via a wavelet transform into multi-resolution subbands: low-frequency components that capture broad, smooth variations and higher-frequency components that encode fine-scale spatial detail. The model estimates the dominant atmospheric contribution primarily from the low-frequency pathway, while skip connections propagate higher-frequency features to reduce oversmoothing and improve preservation of deformation structure in the reconstructed output.

The denoiser operates on a multi-channel input composed of the normalised unwrapped phase together with auxiliary conditioning channels (terrain and water-body descriptors), concatenated along the channel dimension.

\subsection{Preprocessing}
\label{sec:preproc}

Interferograms often contain missing values, usually resulting from poor coherence, water bodies, or the end of acquisition frames. However, the deep learning methods used in this study rely on the spatial or sequential attributes of dense data for effective learning. Therefore, it is necessary to interpolate the data during preprocessing to resemble a dense image. Missing pixels are first filled using nearest‑neighbour interpolation based on surrounding valid values, followed by Gaussian smoothing applied only to the interpolated regions. We use scipy's \texttt{griddata} and \texttt{gaussian\_filter} for interpolation and smoothing. For more details, we refer the reader to our previous work \cite{Popescu2025}.



All interferograms used for training and evaluation are first unwrapped and subsequently normalised to a fixed numerical range to stabilise network training. Let $\phi_{\text{obs}}$ denote the unwrapped interferometric phase. The phase values are scaled by a constant normalisation factor $s_{\phi_{\text{obs}}}$ such that
\begin{equation}
\tilde{\phi}_{\text{obs}} = \frac{\phi_{\text{obs}}}{s_{\phi_{\text{obs}}}},
\end{equation}
where $\tilde{\phi}_{\text{obs}}$ denotes the normalised phase used as network input. The same normalisation is applied to the reference deformation and atmospheric components during training. 

This normalisation ensures that the dynamic range of the input phase remains bounded, improving numerical stability during optimisation and preventing large phase values from dominating the learning process. After inference, the predicted atmospheric phase is rescaled by the same factor to recover the physical phase units. This normalisation also stabilises the estimation of the global phase ramp parameters, whose magnitude depends on the absolute scale of the interferometric phase. After analysing the global dataset, we observe that values outside the range of $[-100, 100]$ rad are rare, so we choose $s_{\phi_{\text{obs}}}=100$ as our regularisation factor.

We additionally derive a binary water mask $m \in \{0,1\}$ (1 = land) from the valid-data mask using morphological cleaning, and compute a signed distance-to-water map
\begin{equation}
d(\mathbf{p}) = \mathrm{EDT}(m(\mathbf{p})) - \mathrm{EDT}(1 - m(\mathbf{p})),
\end{equation}
where $\mathrm{EDT}(\cdot)$ denotes the Euclidean distance transform. We smooth $d$ with a Gaussian kernel ($\sigma = 3$ pixels) to suppress discretisation artefacts, clip to $[-d_{\max}, d_{\max}]$, and normalise to $[-1,1]$. 

The binary mask $m$ provides a hard land/water constraint, while $d$ encodes the distance to water boundaries, enabling the model to account for gradual decorrelation effects in coastal and water-adjacent regions. The network input is formed by concatenating the normalised phase with these two channels $[m, d]$.

To improve robustness to spatial scale and increase training diversity, we adopt a multi-scale random cropping strategy. During training, we sample a crop size
$s_{c} \in \{128,160,192,224,256,288,320,500\}$ pixels (with $s_c=500$ corresponding to the full $500 \times 500$ volcano crop), extract a random $s_c \times s_c$ patch from the preprocessed interferogram, and then resize the patch to a fixed $128 \times 128$ input resolution using bilinear interpolation. The same crop and resizing operation is applied to the DEM conditioning inputs (elevation and its spatial gradients) to preserve pixel-wise alignment between the interferogram and terrain channels. This procedure acts as scale augmentation, allowing the model to learn from both localised and scene-wide atmospheric structure while maintaining a constant network input size for efficient training and inference. For synthetic-data evaluation, we apply the same sampling procedure to ensure consistency with the training distribution. For real-data evaluation on long-term interferograms, we instead resize the full $500\times500$ image to $128\times128$ to obtain a deterministic, full-scene estimation.

Unwrapped interferograms are relative measurements, meaning that the background of the image is not set at zero. To fix this, we take the mean value of a stable reference area and subtract that from the entire image. We use as a reference area an $11\times11$ square around the reference point chosen by the time series generated by LICSBAS \cite{Licsbas}.

\subsection{DINOv3 Cross-Attention (DCA) Conditioning}

To improve the modelling of large-scale spatial dependencies, we incorporate features extracted from a frozen pretrained DINOv3 vision transformer. These features provide hierarchical representations with strong global context, complementing the local modelling capabilities of convolutional networks.

At each WaveDINO encoder, let
$
x \in \mathbb{R}^{B \times C \times H \times W}
$
denote the current feature map and
$
f \in \mathbb{R}^{B \times C_d \times H_d \times W_d}
$
the corresponding DINO feature map. Both feature maps are first normalised using Group Normalisation and then reshaped into token representations
\[
x \rightarrow X \in \mathbb{R}^{B \times N \times C}, \qquad
f \rightarrow F \in \mathbb{R}^{B \times M \times C_d}.
\]

The input $x$ is transformed into the wavelet domain, where its low-frequency component is concatenated with a directly downsampled version of $x$ obtained via PixelUnshuffle, denoted as $x_D$. The LFSS (Low-Frequency State Space) module captures their long-range dependencies via state-space modelling, following a similar design to WaveMamba.

 To incorporate $f$-conditioning, we introduce a DINO cross-attention block (see Fig.~\ref{fig:diagram_big}), where the DINO tokens are projected into the model feature space and reduced through token pooling to limit computational complexity. Cross-attention is then computed using the model tokens as queries and the DINO tokens as keys and values.

To emphasise atmospheric structures, a low-frequency attention bias is introduced by computing a smoothed version of the feature map via average pooling, producing a spatial weighting map $w$ that modulates the attention output
\begin{equation}
\mathrm{Attn}' = \mathrm{Attn} \cdot (1 + w).
\end{equation}

A learnable channel-wise gate regulates the contribution of the attention output before it is added through a residual connection. Finally, the attention-enhanced features are modulated using a FiLM layer \cite{perez2018film} conditioned on the pooled DINO tokens, allowing the transformer features to influence the global statistics of the feature representation. 

\subsection{Terrain-Aware Atmospheric Conditioning}

Atmospheric phase delay in InSAR observations is often correlated with terrain elevation due to stratified tropospheric effects. Variations in elevation influence atmospheric pressure, temperature, and water vapour concentration, producing phase delays that are correlated with topography. To incorporate this physical prior, we introduce a terrain conditioning module that modulates intermediate feature representations using information derived from a digital elevation model (DEM).

\begin{figure}
    \centering
    \if\highres1            
        \includegraphics[width=1\linewidth, trim=0 2mm 0 7mm, clip]{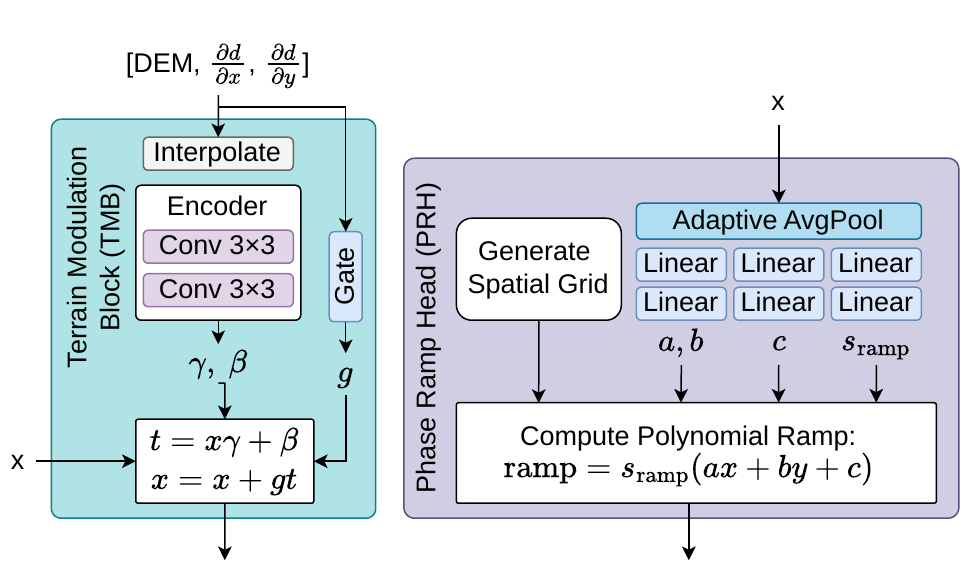}
    \else
        \includegraphics[width=1\linewidth, trim=0 2mm 0 7mm, clip]{images/diagram_2.pdf}
    \fi
    
    \caption{Diagrams of the terrain modulation block (TMB) and the phase ramp head (PRH). TMB conditions feature maps using DEM-derived elevation and gradients, while PRH models global first-order phase ramps from pooled features.}
    \label{fig:diagram_2}
\end{figure}

The proposed Terrain Modulation block (TMB) is presented in \autoref{fig:diagram_2}. In addition to the DEM, we incorporate its spatial gradients ($\frac{\partial d}{\partial x},\frac{\partial d}{\partial y}$) to better capture terrain-driven atmospheric processes.
These gradients encode local slope information that is particularly relevant for orographic atmospheric effects. The terrain input, therefore, consists of three channels: elevation and its horizontal and vertical gradients. Given the feature map $x$ from the low-frequency branch and terrain input $d$ (DEM), the module applies feature-wise modulation:
\begin{equation}
x' = \gamma(d) \odot x + \beta(d),
\end{equation}
where $\gamma(d)$ and $\beta(d)$ are spatially varying scaling and bias terms learned from $d$.

Because atmospheric delay is not always correlated with terrain, particularly in coastal or low-relief regions where turbulence dominates, the terrain conditioning module employs a convolutional gating mechanism to adaptively modulate terrain influence.

After incorporating DEM conditioning, the output is fused with the SKFF (Selective Kernel Feature Fusion) output via the HFE (High-Frequency Enhancement) module. The SKFF module adaptively fuses multi-branch high-frequency features through channel-wise attention guided by global context. The HFE module further enhances high-frequency features using cross-attention and feed-forward modulation conditioned on the low-frequency context (i.e., the output of LFSS). Further details on SKFF and HFE can be found in WaveMamba.

Our decoder blocks follow the design of WaveMamba, where low-frequency components (output of LFSS) are merged with enhanced high-frequency components, followed by an inverse wavelet transform.

\subsection{Phase Ramp Head (PRH)}

Large-scale atmospheric artefacts in unwrapped InSAR frequently contain smooth, low-order spatial components. For the $50$~km crops used in this work, these components are often sufficiently described by an approximately linear trend (a first-order 2D planar ramp) \cite{Murray2021}, although larger scenes may exhibit higher-order curvature and multi-scale structure \cite{Haghshenas2024, Murray2021}. Since convolutional backbones primarily model local interactions, they can be inefficient at capturing global, low-order biases. We therefore introduce a lightweight Phase Ramp Head (PRH), as shown in \autoref{fig:diagram_2}, which predicts a first-order spatial trend from globally pooled decoder features, providing an explicit mechanism to remove dominant ramps while encouraging the remainder of the network to focus on residual structure:
\begin{equation}
\phi_{\mathrm{ramp}}(x,y) \;=\; s_{\mathrm{ramp}}\,\big(a\,x \;+\; b\,y \;+\; c\big),
\label{eq:affine_ramp}
\end{equation}
%
where $x,y \in [-1,1]$ are the normalised horizontal and vertical image coordinates (i.e., $(x,y)\in[-1,1]\times[-1,1]$),
$(a,b,c)\in\mathbb{R}^3$ are bounded coefficients, and $s_{\mathrm{ramp}}\in[-s_{\mathrm{ramp,max}},\, s_{\mathrm{ramp,max}}]$ is a learned global scale that controls the overall ramp magnitude. 

The final atmospheric estimate adds this parametric component to the convolutional output $\phi_{\mathrm{Conv}3\times3}$ as follows.
\begin{equation}
\phi_{\mathrm{atm}} \;=\; \phi_{\mathrm{Conv}3\times3} \;+\; \phi_{\mathrm{ramp}}.
\label{eq:atm_sum}
\end{equation}

In practice, we stabilise training by constraining $(a,b,c)$ and $s_{\mathrm{ramp}}$ via squashing nonlinearities and fixed maxima, which reduces overfitting and prevents the head from absorbing genuine deformation signals. By restricting the parametric head to first‑order trends and explicitly constraining its amplitude, the model captures dominant scene‑wide atmospheric structure while reducing the risk of removing genuine long‑wavelength deformation. 

\begin{table*}
\centering
\caption{Image quality metrics on synthetic data computed over the full image. Metrics include Mean Squared Error (MSE, mm$^2$), Learned Perceptual Image Patch Similarity (LPIPS), Structural Similarity Index (SSIM), $L_3$ norm (mm), and Mean Absolute Error (ABS, mm). The best, second-best, and third-best results in each column are highlighted in red, orange, and yellow, respectively.}
\label{tab:iqm-synthethic}
\begin{tabular}{lcccccccccccc}
\toprule
\multirow{2}{*}{Methods} & \multicolumn{5}{c}{Laguna del Maule} & \multicolumn{5}{c}{Campi Flegrei} \\
\cmidrule(lr){2-6} \cmidrule(lr){7-11}
& MSE$ \downarrow$ & LPIPS $\downarrow$ & SSIM $\uparrow$ & $L_3$$ \downarrow$ & ABS $\downarrow$ 
& MSE $\downarrow$ & LPIPS $\downarrow$ & SSIM $\uparrow$ & $L_3$ $\downarrow$ & ABS $\downarrow$ \\
\midrule
Unet & 54.89 & \cellcolor{third}0.25 & 0.84 & 0.0109 & 4.99 
     & \cellcolor{third}35.30 & \cellcolor{third}0.29 & \cellcolor{third}0.73 & \cellcolor{third}0.0088 & \cellcolor{third}3.37 \\
Diffusion & 93.64 & 0.47 & 0.71 & 0.0140 & 6.58 
          & 66.52 & 0.51 & 0.62 & 0.0117 & 4.48 \\
SwinIR & 69.11 & 0.34 & 0.79 & 0.0125 & 5.74 
       & 40.53 & 0.35 & 0.72 & 0.0097 & 3.83 \\
Restormer & \cellcolor{second}{33.77} & 0.27 & \cellcolor{third}0.85 & \cellcolor{second}0.0090 & \cellcolor{second}3.95 
           & \cellcolor{second}30.42 & 0.46 & 0.64 & \cellcolor{second}0.0081 & \cellcolor{second}3.34 \\
WaveMamba & \cellcolor{third}44.13 & \cellcolor{second}0.16 & \cellcolor{second}0.88 & \cellcolor{third}0.0099 & \cellcolor{third}4.55 
           & 46.29 & \cellcolor{second}0.25 & \cellcolor{second}0.76 & 0.0094 & 3.70 \\
WaveDINO (Ours) & \cellcolor{best}\textbf{30.12} & \cellcolor{best}\textbf{0.12} & \cellcolor{best}\textbf{0.91} & \cellcolor{best}\textbf{0.0080} & \cellcolor{best}\textbf{3.81} 
                & \cellcolor{best}\textbf{26.44} & \cellcolor{best}\textbf{0.09} & \cellcolor{best}\textbf{0.92} & \cellcolor{best}\textbf{0.0063} & \cellcolor{best}\textbf{2.87} \\
\bottomrule
\end{tabular}
\end{table*}

\subsection{Loss function}

Unless otherwise stated, all phase quantities in this section are in the normalised domain (i.e., $\tilde{\phi}=\phi/s_{\phi_{\text{obs}}}$); for brevity, we omit the tilde.

\subsubsection{Atmospheric Estimation Formulation}
The observed unwrapped interferometric phase can be expressed as
\begin{equation}
\phi_{\text{obs}} = \phi_{\text{def}} + \phi_{\text{atm}} + \epsilon ,
\end{equation}
where $\phi_{\text{def}}$ denotes the deformation signal, $\phi_{\text{atm}}$ the atmospheric delay, and $\epsilon$ residual noise and modelling errors.

The network predicts an estimate of the atmospheric component $\hat{\phi}_{\text{atm}}$. The corrected deformation phase is then obtained by subtracting the predicted atmosphere from the observed interferogram:
$
\hat{\phi}_{\text{def}} =
\phi_{\text{obs}} - \hat{\phi}_{\text{atm}}.
$


Training minimises the discrepancy between the corrected deformation $\hat{\phi}_{\text{def}}$ and the synthetic deformation target $\phi_{\text{def}}$ using a masked, reweighted combination of robust and quadratic terms:
\begin{equation}
e = \hat{\phi}_{\text{def}} - \phi_{\text{def}}, \qquad
\mathcal{L}_{\text{def}} =
\big\langle \rho(e)\,\omega \big\rangle_{m}
+\big\langle e^2\,\omega \big\rangle_{m},
\end{equation}
where $\rho(\cdot)$ is the Charbonnier penalty and $\langle X\rangle_m=\frac{\sum mX}{\sum m+\epsilon_0}$ denotes a masked mean over land pixels (water masking the loss). The adaptive weight is
\begin{equation}
\omega = (1 + \lvert \phi_{\text{def}} \rvert)^3 \left( (1 + \lvert e \rvert)^2 - 1 \right) + \epsilon_0 . 
\end{equation}
which emphasises larger deformation magnitudes and harder-to-fit residuals.

Predicting the atmospheric component instead of the deformation signal offers several advantages. Atmospheric phase fields tend to exhibit more consistent spatial statistics across geographic regions, whereas deformation patterns are highly site-specific. Learning the atmospheric component, therefore, improves generalisation across different volcanoes and tectonic environments. Furthermore, this formulation follows the residual-learning paradigm commonly used in image denoising, where the model predicts the noise component rather than the clean signal. Because atmospheric delay typically manifests as a smoother and lower-frequency signal than the deformation signal, directly predicting the atmospheric phase simplifies the learning problem while preserving physical interpretability.

\subsubsection{Ramp Smoothness Regularization}

To ensure that the ramp head models only large-scale atmospheric structure, we introduce a smoothness regularisation term based on the discrete Laplacian of the predicted ramp
\begin{equation}
\begin{split}
\nabla^2 r(x,y) =\;& r(x+1,y) + r(x-1,y) \\
&+ r(x,y+1) + r(x,y-1) - 4r(x,y).
\end{split}
\end{equation}
The corresponding loss term penalises large curvature values
\begin{equation}
\mathcal{L}_{\text{lap}} =
\frac{1}{N}\sum (\nabla^2 r)^2.
\end{equation}
\subsubsection{Final loss}

To preserve deformation structure, we add gradient and local-maxima consistency, a spectral term, and the ramp smoothness regulariser:
\begin{equation}
\mathcal{L}=
\mathcal{L}_{\text{def}}
+\lambda_{\nabla}\mathcal{L}_{\nabla}
+\lambda_{\max}\mathcal{L}_{\max}
+\lambda_{\mathrm{FFT}}\mathcal{L}_{\mathrm{FFT}}
+\lambda_{\mathrm{lap}}\mathcal{L}_{\mathrm{lap}}.
\end{equation}
Here, $\mathcal{L}_{\nabla}$ matches finite-difference gradients of $\hat{\phi}_{\text{def}}$ and $\phi_{\text{def}}$, $\mathcal{L}_{\max}$ matches $9\times 9$ max-pooled responses to preserve local extrema, and $\mathcal{L}_{\mathrm{FFT}}$ encourages agreement of large-scale spectral content. 
In all experiments we use fixed weights $\lambda_{\nabla}=1$, $\lambda_{\max}=1$, $\lambda_{\mathrm{lap}}=0.05$, while $\lambda_{\mathrm{FFT}}$ is increased during training from $0.01$ to $0.1$ for stability. These values were chosen empirically on the validation set and then kept fixed.

\begin{table*}
\centering
\caption{Performance comparison in terms of mean squared error on synthetic data (top) and real data (bottom), evaluated on binned Real deformation. Lower values indicate better performance. The best, second-best, and third-best results in each column are highlighted in red, orange, and yellow, respectively.}
\label{tab:mse_synthetic_real}
\begin{tabular}{lcccccccccc}
\toprule
\multicolumn{11}{c}{Mean squared error on \textbf{Synthetic deformation} in mm} \\
\midrule
Methods & \multicolumn{5}{c}{Laguna del Maule} & \multicolumn{5}{c}{Campi Flegrei} \\
\cmidrule(lr){2-6} \cmidrule(lr){7-11}
Deformation & $<10$ & 10--20 & 20--30 & $>30$ & Mean & $<10$ & 10--20 & 20--30 & $>30$ & Mean \\
\midrule
Unet & 27.76 & 64.39 & 86.76 & 120.02 & 74.73 & 29.92 & 58.57 & 73.39 & 112.79 & 68.67 \\
Diffusion & 39.94 & 89.13 & 138.63 & 247.14 & 128.71 & 63.22 & 103.25 & 132.91 & 204.29 & 125.92 \\
SwinIR & 28.31 & 71.76 & 108.32 & 169.38 & 94.44 & 36.02 & 61.29 & 79.05 & 105.56 & 70.48 \\
Restormer & \cellcolor{best}\textbf{17.81} & \cellcolor{second}35.66 & \cellcolor{second}46.48 & \cellcolor{second}66.82 & \cellcolor{second}41.69 & \cellcolor{third}22.98 & \cellcolor{third}51.29 & \cellcolor{third}73.05 & \cellcolor{third}99.41 & \cellcolor{third}61.68 \\
WaveMamba & \cellcolor{third}22.27 & \cellcolor{third}49.03 & \cellcolor{third}64.47 & \cellcolor{third}86.39 & \cellcolor{third}55.54 & \cellcolor{second}20.88 & \cellcolor{second}41.51 & \cellcolor{second}56.30 & \cellcolor{second}79.91 & \cellcolor{second}49.65 \\
\hline
WaveDINO (Ours) & \cellcolor{second}19.10 & \cellcolor{best}\textbf{35.55} & \cellcolor{best}\textbf{41.80} & \cellcolor{best}\textbf{54.12} & \cellcolor{best}\textbf{37.64} & \cellcolor{best}\textbf{15.62} & \cellcolor{best}\textbf{26.77} & \cellcolor{best}\textbf{37.61} & \cellcolor{best}\textbf{55.29} & \cellcolor{best}\textbf{33.82} \\
\toprule
\multicolumn{11}{c}{Mean squared error on \textbf{Real interferograms} compared to GNSS data in mm} \\
\midrule
Methods & \multicolumn{5}{c}{Laguna del Maule} & \multicolumn{5}{c}{Campi Flegrei} \\
\cmidrule(lr){2-6} \cmidrule(lr){7-11}
Deformation & $<10$ & 10--20 & 20--30 & $>30$ & Mean & $<10$ & 10--20 & 20--30 & $>30$ & Mean \\
\midrule
Unwrapped Interf & 270 & 384 & 404 & \cellcolor{best}\textbf{1484} & \cellcolor{best}\textbf{636} & 167 & \cellcolor{second}125 & \cellcolor{best}\textbf{143} & 448 & 221 \\
GACOS & 393 & 523 & 839 & 3667 & 1355 & 191 & \cellcolor{third}143 & 165 & 464 & 241 \\
Unet & 47 & \cellcolor{second}188 & \cellcolor{second}359 & 3188 & 946 & 136 & 189 & 212 & 640 & 294 \\
Diffusion & 44 & \cellcolor{best}\textbf{182} & 423 & 3169 & 955 & 110 & 206 & 338 & 1495 & 537 \\
SwinIR & \cellcolor{best}\textbf{37} & 205 & 419 & 3360 & 1005 & 113 & 162 & \cellcolor{third}158 & \cellcolor{second}425 & \cellcolor{third}215 \\
Restormer & \cellcolor{second}38 & \cellcolor{third}189 & 397 & 3312 & 984 & \cellcolor{second}90 & 147 & 182 & \cellcolor{third}437 & \cellcolor{second}214 \\
Wavemamba & \cellcolor{third}43 & 193 & \cellcolor{third}365 & \cellcolor{third}2812 & \cellcolor{third}853 & \cellcolor{third}104 & 178 & 175 & 487 & 236 \\
\hline
WaveDINO (Ours) & 48 & 194 & \cellcolor{best}\textbf{355} & \cellcolor{second}2720 & \cellcolor{second}829 & \cellcolor{best}\textbf{42} & \cellcolor{best}\textbf{98} & \cellcolor{second}157 & \cellcolor{best}\textbf{417} & \cellcolor{best}\textbf{173} \\
\bottomrule
\end{tabular}
\end{table*}

\section{Experimental results and discussion}

\subsection{Experimental Setup}
\label{sec:exp_setup}

For training, models are trained per volcano using the mixed \emph{real atmosphere + synthetic deformation} approach, as described in \autoref{sec:synthetic_data}. 
We use \num{38000} synthetic deformation maps, {1647} Laguna del Maule (LdM) interferograms, and {3594} Campi Flegrei (CF) interferograms for training, 
{5588} synthetic maps, {235} LdM and {513} CF interferograms for validation, and {5620} synthetic maps, {470} LdM and {1026} CF interferograms for testing on synthetic data. For testing on real data, we use 2375 Laguna del Maule and 4428 Campi Flegrei interferograms.

We compare our proposed method with (i) unwrapped interferograms, and (ii) GACOS-corrected products, alongside several learned denoisers (UNet, DDPM \cite{jiang2023low}, SwinIR \cite{liang2021swinir}, Restormer \cite{Zamir2022Restormer}, and Wave-Mamba \cite{zou2024wavemamba}) and our model. The state-of-the-art deep learning methods were chosen as representative denoising models covering diverse architectural paradigms, including generative diffusion models, Transformer-based attention mechanisms, and state-space frequency-aware designs.

Training and synthetic-data testing use a multi-scale random crop-and-resize scheme (bilinear resize to 128$\times$128), whereas real-data testing resizes the full 500$\times$500 interferogram (and corresponding DEM channels) directly to 128$\times$128 to produce a deterministic full-scene estimation per interferogram.


\begin{figure*}
    \centering
    \if\highres1            
        \includegraphics[width=1\linewidth]{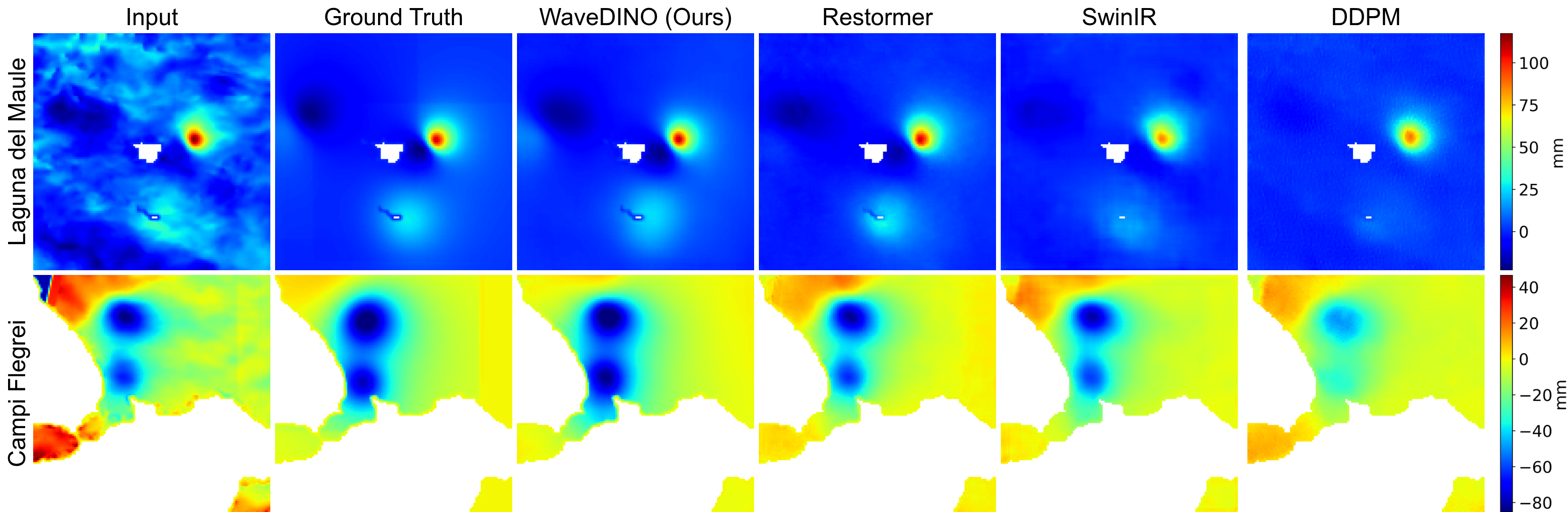}
    \else
        \includegraphics[width=1\linewidth]{images/synthetic_examples.png}
    \fi
    
    \caption{Qualitative comparison on held-out synthetic deformation tests with real atmospheric noise as described in Section \ref{sec:synthetic_data}. Columns show the input, ground truth, and outputs from WaveDINO (ours), Restormer, SwinIR, and DDPM. Top: Laguna del Maule example; bottom: Campi Flegrei example. 
    The white regions indicate masked/low-coherence areas.}
    \label{fig:synthetic_examples}
\end{figure*}


\subsection{Synthetic Deformation Results}

We used two complementary evaluation approaches.
First, we compute standard full-image quality metrics, including Mean Squared Error (MSE), Learned Perceptual Image Patch Similarity (LPIPS), Structural Similarity Index (SSIM), $L_3$ norm (L3), and Mean Absolute Error (ABS). 
However, full-image metrics may be misleading in this setting for two reasons: (i) large regions of the ground-truth deformation are close to zero, which can dominate global error measures (e.g., MSE) and mask performance on the deforming areas; and (ii) evaluating over the entire frame primarily quantifies atmospheric removal, without directly testing whether the deformation signal is preserved. 
Therefore, our second approach focuses explicitly on deformation preservation by sampling 100 pixels per image and binning them by the ground-truth deformation magnitude ($<10$\,mm, 10--20\,mm, 20--30\,mm, and $>30$\,mm), after which we compute MSE within each bin. This stratified evaluation also provides an additional benefit: it reveals magnitude-dependent behaviour (e.g., different performance on subtle versus large deformation), which may be obscured by a single global score.

Table~\ref{tab:iqm-synthethic} reports the whole-image quality metrics on the held-out synthetic test set, where synthetic deformation is added on real interferograms that exhibited stable ground. These metrics quantify overall restoration fidelity, but they are dominated by large regions with near-zero deformation and therefore primarily reflect the ability to suppress atmospheric structure rather than preserve deformation. Under this setting, WaveDINO achieves the best (or second-best) perceptual and structural scores across both volcanoes (lowest LPIPS and highest SSIM), indicating improved atmospheric attenuation without excessive smoothing.

To directly evaluate deformation preservation, we compute MSE within bins of ground-truth deformation magnitude (Table~\ref{tab:mse_synthetic_real} top). This stratified analysis reveals magnitude-dependent behaviour that is obscured by global metrics. At Campi Flegrei, WaveDINO yields the lowest error across all deformation bins and achieves the best mean error, demonstrating improved recovery of subtle signals ($<10$ mm) as well as larger displacement. At Laguna del Maule, performance varies by bins: transformer-based baselines perform well for small deformation, whereas WaveDINO remains competitive and improves robustness at larger deformation magnitudes, supporting its ability to preserve deformation structure while suppressing realistic atmospheric contamination. 

Figure~\ref{fig:synthetic_examples} illustrates representative synthetic test cases. Compared to diffusion and SwinIR, WaveDINO better preserves the spatial footprint and gradients of the deformation signal while reducing broad atmospheric structure. Restormer often preserves deformation morphology but can leave residual low-frequency artefacts in some cases, consistent with its mixed quantitative performance across the two sites.

\subsection{Real Deformation Results}

We evaluate deformation accuracy on real images against GNSS by reporting the MSE (in \si{mm}) binned by deformation magnitude, matching the GNSS time window to the SAR acquisition dates. The results are averaged across all available GNSS stations in each study area (five at Laguna del Maule and 19 at Campi Flegrei after quality filtering). To ensure a comprehensive and unbiased evaluation, we use a large set of long-term interferograms (up to 5,000 per volcano), avoiding subjective selection and including intervals with weak deformation. This provides a realistic assessment of both deformation recovery and false-positive behaviour.

Table~\ref{tab:mse_synthetic_real} (bottom)  compares the results against GNSS by matching the GNSS displacement window to each interferogram's acquisition dates, projecting to the same LOS component, and computing binned MSE by deformation magnitude. 
The results show that learned denoising substantially improves agreement with independent GNSS compared with raw unwrapped interferograms and those corrected by GACOS. At Campi Flegrei, WaveDINO achieves the lowest mean MSE among all the methods, whereas GACOS can produce large errors in some intervals, reflecting sensitivity to atmospheric conditions and model limitations. At Laguna del Maule, most learning-based methods improve over the uncorrected interferograms for deformation values below 30mm. WaveDINO yields the best mean error out of all the learning-based models and reduces error, particularly in the highest deformation bin relative to other learning-based baselines, indicating improved preservation of large-amplitude deformation while suppressing atmospheric structure.

\begin{figure}
    \centering
    \if\highres1            
        \includegraphics[width=1\linewidth]{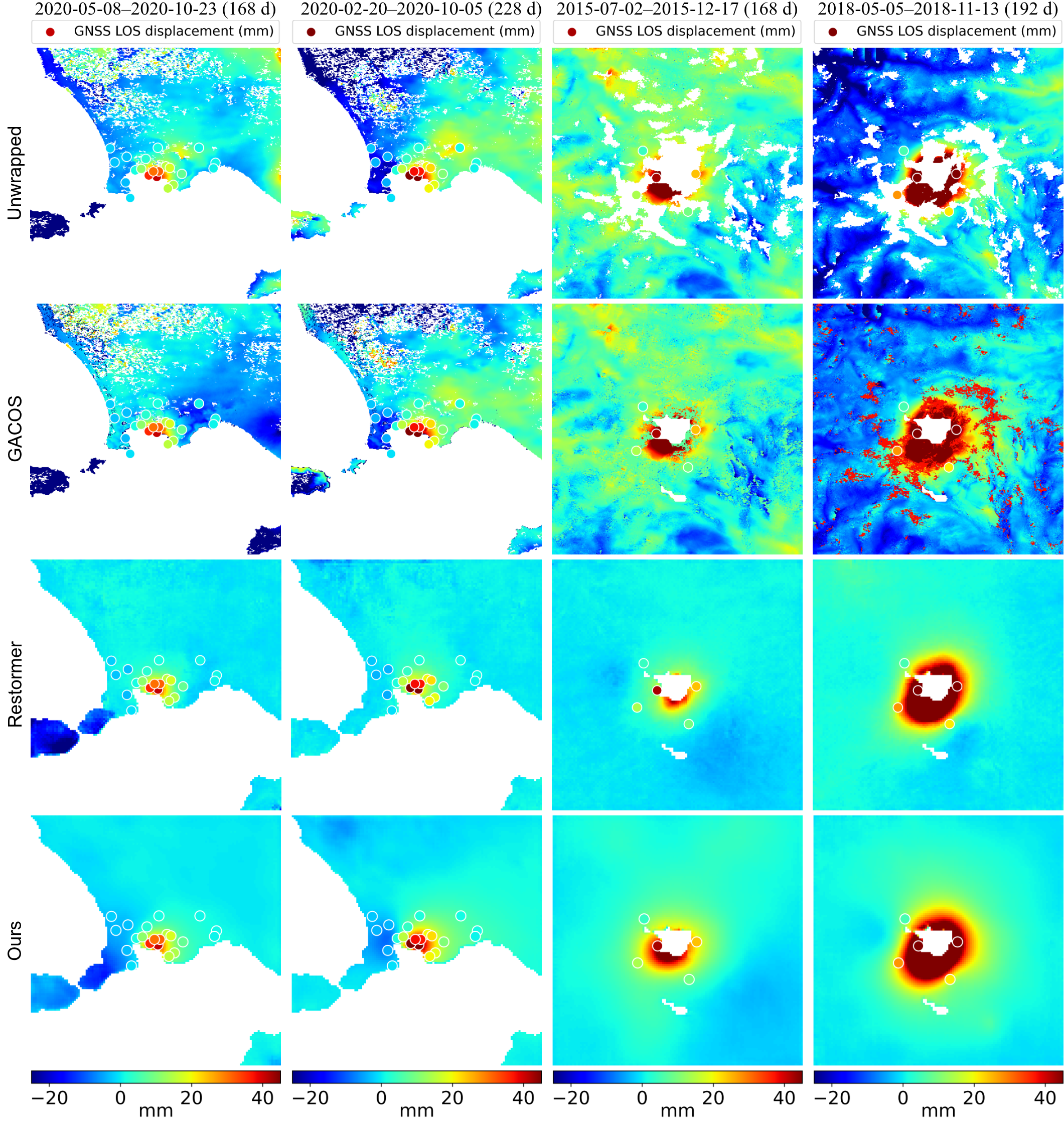}
    \else
        \includegraphics[width=1\linewidth]{images/campi_examples.png}
    \fi
    
    \caption{Qualitative comparison on long-term interferograms from Campi Flegrei ($40.827^\circ$N, $14.139^\circ$E; first two columns) and Laguna del Maule ($36.058^\circ$S, $70.492^\circ$W; last two columns). 
    Rows show (top to bottom) the unwrapped interferogram, GACOS correction, WaveDINO (ours), and Restormer. 
    GNSS stations are overlaid, coloured by accumulated LOS displacement (mm) with white outlines for visibility. 
    Each panel is approximately 50~km across.
    }
    \label{fig:examples_campi}
\end{figure}

Figure~\ref{fig:examples_campi} highlights the behaviour of the different methods across two contrasting volcanic environments. At Campi Flegrei, atmospheric artefacts are spatially heterogeneous and influenced by coastal and urban atmospheric variability, while at Laguna del Maule, they are dominated by smooth, large-scale phase delays correlated with strong topographic relief and high-altitude meteorological conditions. GACOS reduces some broad-scale structure at both sites but leaves substantial residual artefacts, particularly at Laguna del Maule. Both learning-based methods suppress atmospheric contamination more effectively; however, WaveDINO more consistently removes scene-wide gradients while preserving coherent, localised deformation signals across both volcanoes, yielding deformation patterns that align more closely with independent GNSS observations.

The remaining discrepancies are dominated by factors that are not purely atmospheric: coherence loss (e.g., seasonal snow at Laguna del Maule) introduces missing-data regions that must be inpainted and can bias both learned and model-based corrections. GNSS time series also contain gaps and noise, resulting in interpolation that introduces uncertainty into the reference values used for evaluation. Despite these limitations, the GNSS comparison indicates that hybrid training (synthetic deformation + real atmospheric noise) produces models that better preserve deformation than weather-model corrections in these case studies, supporting their use as a complementary tool for operational volcano monitoring.

\subsection{Ablation Study}

\begin{table}[t]
\centering
\caption{Ablation study of the proposed framework on Laguna del Maule on synthetic data.}
\label{tab:ablation}
\begin{tabular}{lcccccc}
\hline
Variant & Water & DEM & Ramp & DINO & MSE $\downarrow$ & \\
\hline
V1 & $\times$ & $\checkmark$ & $\checkmark$ & $\checkmark$ & \cellcolor{third}33.51 \\
V2 & $\checkmark$ & $\times$ & $\checkmark$ & $\checkmark$ & 37.57  \\
V3 & $\checkmark$ & $\checkmark$ & $\times$ & $\checkmark$ & 34.08 \\
V4 & $\checkmark$ & $\checkmark$ & $\checkmark$ & $\times$ & \cellcolor{second}32.39  \\
Ours & $\checkmark$ & $\checkmark$ & $\checkmark$ & $\checkmark$ & \cellcolor{best}\textbf{30.12}  \\
\hline
\end{tabular}
\end{table}


Table~\ref{tab:ablation} shows that removing any component degrades performance. DEM conditioning is most influential (V2: MSE 37.57 vs. 30.12), followed by ramp removal (V3: 34.08), water-feature removal (V1: 34.51) and DINO removal (V4: 32.39). The full model performs best (MSE 30.12), indicating complementary benefits from terrain, water-aware inputs, ramp modelling, and DINO features.

\section{Conclusion}

In this study, we introduced WaveDINO for learning-based atmospheric correction of unwrapped InSAR interferograms, designed to suppress tropospheric artefacts while preserving volcanic deformation. Across synthetic tests and long-term interferograms at Laguna del Maule and Campi Flegrei, WaveDINO reduces atmospheric structure and improves agreement with independent GNSS compared with uncorrected and weather-model-based corrections. Future work will extend the approach to larger spatial domains and incorporate temporal context to further improve generalisation.



\section*{Data and Code Availability}
The LiCSAR Sentinel-1 interferograms for Laguna del Maule and Campi Flegrei are discoverable via the COMET Volcano Portal and LiCSAR archives at \textbf{\url{http://comet.nerc.ac.uk}}. 
The synthetic deformation generator is available at \textbf{\url{https://github.com/pui-nantheera/Synthetic_InSAR_image}}.
Code for preprocessing, training, inference, and experiment configurations will be released upon acceptance at \textbf{\url{https://github.com/robertpop99/WaveDINO}}. 
We use a frozen DINOv3 ConvNeXt-Base checkpoint (\texttt{dinov3-convnext-base-pretrain-lvd1689m}).

\section*{Acknowledgment}
The authors thank L. C\'ordova and the Observatorio Volcanol\'ogico de los Andes del Sur (OVDAS), SERNAGEOMIN (Temuco, Chile), for providing daily GNSS time series from Laguna del Maule (stations COLO, LDMP, MAU2, NIEB, and PUEL) covering 2014-01-01 to 2023-12-30.

\small
\bibliographystyle{IEEEtran}
\bibliography{egbib}


 
 

\end{document}